\documentclass[12pt,epsfig]{article}
\usepackage{amsmath}
\usepackage{graphicx}
\usepackage{epstopdf}
\usepackage{lscape}
\usepackage{longtable}
\usepackage[usenames, dvipsnames]{color}
\begin{document}
	\begin{center}
		\textbf {A new nonparametric interpoint distance-based measure for assessment of clustering}\\
	\end{center}
		\begin{center}
		Dr. Soumita Modak$^{*}$\\
		Faculty of Statistics\\
		University of Calcutta\\
		Basanti Devi College\\
		147B, Rash Behari Ave, Kolkata- 700029, India\\
		Email: soumitamodak2013@gmail.com\\
		Orcid id: 0000-0002-4919-143X\\
		Homepage: https://sites.google.com/view/soumitamodak
	\end{center}
	Abstract: A new interpoint distance-based measure is proposed to identify the optimal number of clusters present in a data set. Designed in nonparametric approach, it is independent of the distribution of given data. Interpoint distances between the data members make our cluster validity index applicable to univariate and multivariate data measured on arbitrary scales, or having observations in any dimensional space where the number of study variables can be even larger than the sample size. Our proposed criterion is compatible with any clustering algorithm, and can be used to determine the unknown number of clusters or to assess the quality of the resulting clusters for a data set. Demonstration through synthetic and real-life data establishes its superiority over the well-known clustering accuracy measures of the literature.   
	
	keyword: Clustering accuracy measure, Interpoint distance, Nonparametric statistic, Spearman's rank correlation coefficient, High-dimensional applicability. 
	\section{Introduction}
	In the search for physical origins behind the data these days scientists are dealing with challenging big sets of data through efficient algorithms using the help of computers. However, it is easier to study their properties when data sets can be split into meaningful classes.
	This kind of study includes unsupervised classifications where we try to find out such classes by placing the closer members of the data set in the same group or cluster and further members in different groups, it is called a cluster analysis. We consider the situations where all the members are classified into mutually exclusive and exhaustive clusters. There are different kinds of clustering methods namely, partitioning, hierarchical, model-based, grid-based and density-based (Jain et al. 1999; McLachlan \& Peel 2000; Kaufman \& Rousseeuw 2005; Cheng et al. 2017, 2018, 2021; Matioli et al. 2018; Modak et al. 2018, 2020, 2021; Tarnopolski 2019; Toth et al. 2019; Modak 2019). By construction or assumptions involved in the clustering methods, choice of the algorithm is problem specific. For example, Gaussian model-based or $K-$means clustering algorithms are known to work quite well for spherical clusters, whereas agglomerative hierarchical clustering with average linkage is preferred for ball-shaped clusters. Nevertheless, however appropriate clustering method is chosen, validation of the clustering output is a crucial task for revealing the true clusters present in the given data set. It is to noted that even if a plausible clustering algorithm is implemented, the unknown true value for the number of clusters $(K)$ can only be revealed by a proper validity index. In fact, for a given data set the same algorithm could imply different values for $K$ in terms of various independent validity measures. 
	
	Parametric mixture models give rise to clustering accuracy measures like the classic Bayes'
	Information Criterion and Bayes factor (see, for example, Schwarz 1978; Kass and Raftery 1995; Frayley and Raftery 1998; Toth et al. 2019). Sugar and James (2003) suggest the jump method to extract the true number of groups from a given data set by applying a parametric model-based approach using the Mahalanobis distance. Tibshirani et al. (2001) propose the `gap statistic' to estimate the number of clusters present in a set of data. Their method has usability in association with any clustering algorithm and distance measure. However, it requires a reference distribution to be specified appropriately for the given data, which is not only difficult to choose but also computationally extensive. Distance-based nonparametric measures include the widely used Dunn index (Dunn 1974), connectivity (Handl et al. 2005), Cali\'{n}ski and Harabasz index (Cali\'{n}ski \&  Harabasz 1974), nearest neighbor classification error rate (Ripley 1996), and the cluster assessment index proposed by Modak (Modak 2022).    
	Another such popular measure is the average silhouette width (Rousseeuw 1987) which is improved by Cheng et al. (2019) with a new cluster validity index, based on local cores, which is effective for determining arbitrarily shaped clusters.
	
	In this paper, we propose a novel accuracy measure which is capable of determining the unknown number of clusters existing in a given set of data. On the other hand, for a specific number of groups it assesses the quality of a cluster analysis performed as well as compares the performances of different clustering algorithms for a particular data set. The suggested assessment measure is compatible with any clustering algorithm to be implemented. Our cluster validity index is based on the interpoint or intermember distances of the data set, wherein the two terms `member' and its corresponding `observation' are used interchangeably to compute the distance measure (e.g., suppose we are clustering ten stars using their observations on the variables brightness and mass, then the interpoint distance between two star members is the computed value for the distance between their corresponding observation vectors in the brightness-mass space). 
	Our method is flexible enough to work with any distance measure (may not be strictly metric) appropriate for the given data. This property makes our approach applicable to data measured on arbitrary scales (e.g., with  data given on a ratio scale one may use the Euclidean norm, whereas for mixed data measured on both ratio and nominal scales the Gower's distance may be considered, for details, see, Kaufman and Rousseeuw 2005). The proposed method is nonparametric which does not require any model assumption regarding the data and hence is independent of the distribution of data under study. Interpoint distance makes the method feasible in any dimensional space, i.e. data can be univariate, multivariate or even high-dimensional where the number of observations can be close to or less than the number of study variables. We design our measure in such a simple fashion that it is convenient to compute and easy to interpret. Variables constructed in a sophisticated way are finally put in the form of the sample version of a popular nonparametric statistic, that is, the Spearman's rank correlation coefficient. We suggest our assessment index in terms of the arithmetic mean of such Spearman's rank correlation coefficients computed for all the members in the data. It lies from -1 to 1, where a larger value indicates a better cluster analysis and its computed value with respect to the boundaries gives a vivid understanding of how well or bad a classification is.
	
	We exhibit the ability of our measure to recognize the natural clusters in the data sets through simulated groups. We use closely spaced spherical clusters and overlapping groups of arbitrary shapes with noisy observations. We also consider high-dimensional data under complex dependence structure defined by copula (Nelsen 2006; Modak \& Bandyopadhyay 2019), where the number of study variables is higher than the sample size. Our assessment index finds the groups in real data sets concerning (i) worldwide $CO_2$ emission (Matioli et al. 2018), (ii) `CYG OB1' star cluster (Vanisma \& Greve 1972; Kaufman \& Rousseeuw 2005), and (iii) Universe's brightest source of light, i.e. gamma-ray bursts (Bandyopadhyay \& Modak 2018; Modak et al. 2018; Modak 2021). In this paper, we carry out the widely used $K-$means (Hartigan \& Wong 1979), hierarchical (Kaufman \& Rousseeuw 2005) and DBSCAN clusterings (Ester et al. 1996; Campello et al. 2013), and the Euclidean metric is used as the distance measure. Our proposed criterion proves itself competitive with the superiority in most of the situations compared to the other distance-based accuracy measures  under consideration, namely the Dunn index (Dunn 1974), connectivity (Handl et al. 2005), Cali\'{n}ski and Harabasz index (Cali\'{n}ski \&  Harabasz 1974), and nearest neighbor classification error rate (Ripley 1996).    
	
	The paper is so designed that Section 2 describes our proposed method in details. Section 3 briefly discusses the competitors considered in the present study, and demonstrates the usefulness of our measure through simulated and real-life data sets. In Section 4, the conclusion is drawn.
	
	\section{Method}\label{method}
	To assess the quality of a clustering algorithm we need to analyze whether the close members are clustered in the same group and far members classified in different groups. In real-life situations data sets mostly happen to be so complex that the resulting clusters of data members are very closely related or even overlapping with respect to the study variables to which the clustering is applied. Therefore, we investigate whether the member is more likely to fall in the group it has been clustered in compared to the closest neighboring group. Here closeness among the members is measured in terms of the interpoint distances and the nearest cluster of a member is defined as the cluster to which the member is not classified during the cluster analysis but has the minimum average distance with all its members among all the clusters it has not been classified to. 
	
	We assume that the given data set has more than one inherent group in it (i.e. $K>1$) and concentrate on finding the unknown true value of $K$.
	Suppose the data set of size $n$ is clustered using a clustering algorithm into $K(>1)$ mutually exclusive and exhausted clusters $C_1$,\ldots,$C_K$ of sizes $n_1,\ldots,n_K$ respectively. Let $M_{k,m}$ denote the $m-$th member of the $k-$th cluster $C_k$ or the corresponding observation in the data set for $m=1,\ldots,n_k$ and $k=1,\ldots,K$. The distance between any two members $x$ and $y$ of the data is represented by $d(x,y)$. Construction of our clustering assessment index is discussed below.\\ 
	(a) Consider the $m-$th member of cluster $C_k$, i.e. $M_{k,m}$. \\ 
	(a1) Compute its average distance to all members for each of the other clusters as: 
	\begin{equation*}
		d_{k',m}=\frac{1}{n_{k'}}\sum\limits_{m'=1}^{n_{k'}}d(M_{k,m},M_{k',m'})\hspace{.1 in} \text{for}\hspace{.1 in} k'(\neq k)=1,\ldots,K.
	\end{equation*}
	(a2) We find the nearest cluster of the member, denoted by $C_{nc}$, for which
	\begin{equation}\label{NC}
		\underset{1\leq k'(\neq k)\leq K}{\min}\hspace{.05 in}\bigg\{d_{k',m}\bigg\}=d_{nc,m}\hspace{.1 in}\text{holds}.
	\end{equation} 
	If Eq.~\eqref{NC} is attained numerically for more than one cluster, then the nearest cluster can be chosen at random from them.\\
	(b) Now we consider only two sets of distances.\\
	(b1) Distances of the member to all other members of its own cluster $C_{k}$, i.e.
	\begin{equation*}
		d(M_{k,m},M_{k,m'})\hspace{.1 in} \text{for}\hspace{.1 in} m'(\neq m)=1,\ldots,n_{k}.
	\end{equation*} 
	(b2) Distances of the member to all members of the nearest cluster $C_{nc}$, i.e. 
	\begin{equation*}
		d(M_{k,m},M_{nc,m'})\hspace{.1 in} \text{for}\hspace{.1 in} m'=1,\ldots,n_{nc}.
	\end{equation*} 
	(b3) Compute the maximum of the two sets of distances from (b1) and (b2), let it be denoted by $M$, i.e.
	\begin{equation*}
		M=\max\hspace{.05 in}\bigg\{\underset{m'(\neq m)=1,\ldots,n_{k}}{\max}\hspace{.05 in}\{d(M_{k,m},M_{k,m'})\},\hspace{.05 in}\underset{m'=1,\ldots,n_{nc}}{\max}\hspace{.05 in}\{d(M_{k,m},M_{nc,m'})\}\bigg\},
	\end{equation*}
	which is always greater than zero.\\
	(b4) Normalize the distances from (b1) and (b2) by dividing them by $M$ so that all the distances belong to the set $(0,1]$. \\
	(c) Now, our final two working sets of distances are as follows:\\
	(c1) First set:\\
	\begin{equation*}
		\frac{1}{M}\hspace{.05 in}d(M_{k,m},M_{k,m'})\hspace{.1 in} \text{for}\hspace{.1 in} m'(\neq m)=1,\ldots,n_{k}.
	\end{equation*} 
	(c2) Second set:\\
	\begin{equation*}
		\frac{1}{M}\hspace{.05 in}d(M_{k,m},M_{nc,m'})\hspace{.1 in} \text{for}\hspace{.1 in} m'=1,\ldots,n_{nc}.
	\end{equation*} 
	(c3) We divide the set $(0,1]$ into $N$ number of equally spaced, mutually exclusive and exhaustive intervals of width $w\in(0,1)$ as:\\ $(0,w],(w,2w],\ldots,(\{N-1\}\times w,N\times w]$ with $N\times w=1$.\\
	(c4) We compute the numbers of distances from (c1) and (c2) separately which fall in the intervals defined in (c3) such that\\
	(c4.1) $f_1(x)$: frequency of (c1) distances lying in an interval  $(x-\frac{w}{2},x+\frac{w}{2}]$ and\\
	(c4.2) $f_2(x)$: frequency of (c2) distances lying in an interval $(x-\frac{w}{2},x+\frac{w}{2}]$,\\
	where $x$ is the midpoint of an interval.\\  
	(d) Suppose $x_h=$ midpoint of the $h-$th interval $(\{h-1\}\times w,h\times w], h=1,\ldots,N$, and the corresponding frequencies of the distances from (c1) and (c2) are respectively $f_1(x_h)$ and $f_2(x_h)$.\\
	(d1) We rank the two series $\{f_2(x_h)-f_1(x_h),h=1,\ldots,N\}$ and $\{x_h,h=1,\ldots,N\}$ independently. Suppose that $R_{1h}$ denotes the rank of $[f_2(x_h)-f_1(x_h)]$ among $[f_2(x_1)-f_1(x_1)],\ldots,[f_2(x_N)-f_1(x_N)]$. Let $R_{2h}$ represent the rank of $x_h$ among $x_1,\ldots,x_N$, with clearly $\{R_{2h},h=1,\dots,N\}=(1,\ldots,N)'$ as $x_1<\ldots<x_N$.\\
	(d2) Compute the following statistic:\\
	\begin{equation}\label{spearman}
		R_{k,m}=\frac{12\sum\limits_{h=1}^{N}\bigg\{\big(R_{1h}-\frac{N+1}{2}\big)\big(R_{2h}-\frac{N+1}{2}\big)\bigg\}}{N(N^2-1)},
	\end{equation}
	which is the sample version of the Spearman's rank correlation coefficient between the two dependent variables $[f_2(x)-f_1(x)]$ and $x$. Eq.~\eqref{spearman} gives the expression for no tie case, which can be adapted in the presence of ties among the $\{f_2(x_h)-f_1(x_h),h=1,\ldots,N\}$ observations by giving rank to each of the observations in a tied group equal to the average of
	integer ranks corresponding to that tied group.\\
	
	(e) Repeat the above stated operations for each member of every cluster, i.e. for every $M_{k,m}$ with $m=1,\ldots,n_k$ and $k=1,\ldots,K$. Then we define our proposed statistic R$_{clus}$ to measure the accuracy of a cluster analysis as follows:
	\begin{equation}\label{measure}
		\text{R}_{clus}=\frac{1}{n}\sum\limits_{k=1}^{K}\sum\limits_{m=1}^{n_k}R_{k,m}\hspace{.01in},
	\end{equation}
	where $n=\sum\limits_{k=1}^{K}n_k$.
	Clearly, $-1\leq \text{R}_{clus}\leq 1$ with greater value indicating better clustering.
	\subsection{Analysis of the proposed method}
	For each specified member in the data set $M_{k,m}$, $S_1=\{d(M_{k,m},M_{k,1}),\\d(M_{k,m},M_{k,2}),\ldots,d(M_{k,m},M_{k,\overline{m-1}}),
	d(M_{k,m},M_{k,\overline{m+1}}),\ldots,d(M_{k,m},M_{k,n_k})\}$ and\\ $S_2=\{d(M_{k,m},M_{nc,1}),d(M_{k,m},M_{nc,2}),\ldots,d(M_{k,m},M_{nc,n_{nc}})\}$ denote two independent sets of interpoint distances of $M_{k,m}$ to members of its own cluster $C_k$ and its nearest cluster $C_{nc}$ respectively . Then accuracy of clustering the member $M_{k,m}$ in the cluster $C_k$ can be assessed in the following manner:\\
	(i) If the member is correctly clustered in $C_k$, then the distances from the set $S_2$ tend to be larger than the distances of the set $S_1$.\\
	(ii) If the member is incorrectly clustered in $C_k$, then the distances from the set $S_2$ tend to be smaller than the distances of the set $S_1$.\\
	(iii) If the member is equally probable to be clustered in any of the clusters $C_k$ and $C_{nc}$, then the distances of the two sets $S_1$ and $S_2$ tend to be close.\\  
	The same ratiocination is valid for the interpoint distances belonging to the sets $S'_1=S_1/M$ and $S'_2=S_2/M$, where all the distances take values in $(0,1]$.
	
	Under situation (i), as we move from 0 to 1 along the set $(0,1]$, we expect an overall decreasing pattern in the number of distances from the set $S'_1$, i.e. $f_1(x) \downarrow$ in $x$, and simultaneously we expect a general increase in the number of distances from the set $S'_2$, i.e. $f_2(x) \uparrow$ in $x$. Equivalently, we can say $[f_2(x)-f_1(x)] \uparrow$ in $x$. In such situation, the computed value for the sample Spearman's rank correlation coefficient $R_{k,m}$ should be close to 1. Likewise, the condition (ii) brings about the reverse, i.e. we anticipate $f_1(x) \uparrow$ in $x$ and $f_2(x) \downarrow$ in $x$ $\Leftrightarrow$ $[f_2(x)-f_1(x)] \downarrow$ in $x$ and we should get a value close to -1 for $R_{k,m}$. For the case (iii), a value close to 0 is expected, where both $f_1(x)$ and $f_2(x)$ are supposed to have similar kind of trends with respect to $x$ and hence the trend of $[f_2(x)-f_1(x)]$ is likely to be non-monotonic function of $x$ without showing any increasing or decreasing pattern in $x$. Now, we compute $R_{k,m}$ for all members of the data set and consider their arithmetic mean as our proposed measure R$_{clus}$. Thus, it is a nonparametric statistic with $-1\leq \text{R}_{clus}\leq 1$ whose larger value indicates better clustering. 
	
	In this context, it is interesting to give a comparative overview of our measure and the popular nonparametric clustering accuracy measure the average silhouette width which is defined as follows: for each member $M_{k,m}$, let
		$a(M_{k,m})$ be the average of interpoint distances of $M_{k,m}$ to all other members
		of its own cluster and $b(M_{k,m})$ be the minimum of average interpoint distances of $M_{k,m}$ to members of other clusters. Then, the silhouette width for the member $M_{k,m}$, $s(M_{k,m})$ is defined as
		\begin{equation}
			s(M_{k,m})=\frac{b(M_{k,m})-a(M_{k,m})}{\max\bigg\{a(M_{k,m}),b(M_{k,m})\bigg\}}.
		\end{equation}
		The average silhouette width is (Rousseeuw 1987; Kaufman and\\ Rousseeuw 2005)
		\begin{equation}
			\text{ASW}=\frac{1}{n}\sum\limits_{k=1}^{K}\sum\limits_{m=1}^{n_k}s(M_{k,m}),
		\end{equation}
		ASW $\in [-1,1]$ whose larger value suggests better classification.
		The average silhouette width and our proposed index R$_{clus}$ both are designed in the form of some simple arithmetic means of the respective measures, where the measure is the silhouette width in the case of the average silhouette width, and in our case the measure is the correlation coefficient. Values of these two indices lie from -1 to 1 where higher values indicate better clustering. Our assessment index is designed in this manner to make the numerical results easily interpretable and it is achieved by using the property of the Spearman's rank correlation coefficient that its value always ranges from -1 to 1, and so is true for the silhouette width by its construction. Moreover, both of the measures rely on the comparison between the nearest cluster and the assigned cluster of each member but in completely different ways. Our method deals with two sets $S_1$ and $S_2$ and develops a sophisticated novel measure using these sets, where the Spearman's rank correlation coefficient is computed between our designed variables $[f_2(x)-f_1(x)]$ and $x$ for all members, whereas the silhouette width considers simply $a(M_{k,m})$ from the assigned cluster and $b(M_{k,m})$ from the nearest cluster for the member $M_{k,m}$ for each $k$ and $m$.
	
	In our measure, it is to be noted that each $R_{k,m}$ and consequently R$_{clus}$ are functions of $N$ (or $w$), which is a data-dependent choice. Ideally, we should use the notations $R_{k,m}(N)$ (or $R_{k,m}(w)$) for all $k,m$ and R$_{clus}(N)$ (or R$_{clus}(w)$), where $N$ (or $w$) takes a specified value. However, for simplicity, we suppress $N$ from the notations. For different values of $N$ or $w$, the computed values of $R_{k,m}$ s and hence R$_{clus}$ differ. Therefore, our measure should be calculated for the various values of $w$ and the results are to be accepted when they are robust regarding all the considered values (it is illustrated in the third simulation under Section 3.1). As we need to compute the Spearman's rank correlation coefficient for all the members (see, Eq.~\eqref{spearman}) to obtain our required index (as in the Eq.~\eqref{measure}), naturally with increasing sample size the computational steps increase.
	\subsection{Illustration of the method through an example}
	Suppose without loss of generality, $k=1$ and $m=1$, then $M_{k,m}$ denotes the first member of the first cluster attained. As stated under step (a), among all clusters except the first one, let the second cluster be the one whose average distance from $M_{1,1}$ is a minimum, then the second cluster is defined as the nearest cluster of $M_{1,1}$ which is denoted by $C_{nc}$ for $nc=2$. Distances between $M_{1,1}$ and all other members of the first cluster (from step (b1)) are divided by $M$ (as found in Step (b3)) and we obtain the observations defined in step (c1). Similarly, distances between $M_{1,1}$ and all members belonging to the second cluster (from step (b2)) are divided by $M$ and we have the observations defined in step (c2). Division by $M$ makes all the distances from (c1) and (c2) fall in the interval $(0,1]$ which is divided into, as described in the step (c3), $(0,0.1],(0.1,0.2],\ldots,(0.9,1]$ for $N=10$ (i.e. $w=0.1$). Now count the numbers of (c1) and (c2) distances separately lying in each of the sub-intervals (as explained in the steps (c4.1) and (c4.2)). $f_1(0.05)$ denotes the number of (c1) distances lying in the sub-interval $(0,0.1]$ and $f_2(0.05)$ represents the number of (c2) distances falling in $(0,0.1]$, similarly $f_1(0.15)$ and $f_2(0.15)$ are for the sub-interval $(0.1,0.2]$, and so on. Thus, as defined in step (d), we have two series $\{f_2(x_h)-f_1(x_h),h=1,2,\ldots,10\}$ and $\{x_h,h=1,2,\ldots,10\}$, where $x_1=0.05,x_2=0.15,\ldots,x_{10}=0.95$. Subsequently, $R_{1h}$ denotes the rank of $[f_2(x_h)-f_1(x_h)]$ among $[f_2(0.05)-f_1(0.05)],[f_2(0.15)-f_1(0.15)],\ldots,[f_2(0.95)-f_1(0.95)]$ and $R_{2h}$ denotes the rank of $x_h$ among $0.05,0.15,\ldots,0.95$, i.e. $\{R_{2h},h=1,2,\dots,10\}=(1,2,\ldots,10)'$.
	
	If the member $M_{1,1}$ originally belongs to the first cluster, then its interpoint distances from all other members of first cluster are on an average smaller than those from all members of the second cluster (i.e. the nearest cluster of $M_{1,1}$). As a result, we expect more or less a downward trend in the series $\{f_1(0.05),f_1(0.15),\ldots,f_1(0.95)\}$ and simultaneously an upward trend in $\{f_2(0.05),f_2(0.15),\ldots,f_2(0.95)\}$, or in other words, an increasing pattern in $\{f_2(0.05)-f_1(0.05),f_2(0.15)-f_1(0.15),\ldots,f_2(0.95)-f_1(0.95)\}$ as we move along from left to right over the increasing sub-intervals $(0,0.1],(0.1,0.2]\ldots,(0.9,1]$ or equivalently over the corresponding mid-points of the sub-intervals $\{x_1=0.05,x_2=0.15,\ldots,x_{10}=0.95\}$. This produces a high positive value of Spearman's rank correlation coefficient between the two variables $[f_2(x)-f_1(x)]$ and $x$ (see, step (d2)). Likewise, a high negative value can be connected to the situation that the member is more likely to be part of its nearest cluster. On the other hand, a value close to 0 indicates that the member is equally likely to be a member of both the clusters. This concept is repeated for all $k$ and $m$ and finally we compute the simple arithmetic mean of such Spearman's rank correlation coefficients computed for all the members of the data set (step (e)).
	\section{Numerical experiments}
	In this study, we choose $N=10$, i.e. $w=0.1$ unless mentioned otherwise.
	We consider the following four other distance-based validity indices, widely used for assessment of a classification, to understand the relative performance of our proposed measure.\\ 
	
	Dunn index: The Dunn index is defined as the ratio of the smallest distance between members of different clusters to the largest intra-cluster distance (Dunn 1974). It is expressed as
	\begin{equation}
		\text{Dunn}=\frac{\underset{k\neq k'}{\underset{1\leq k,k' \leq K}{\min}}\bigg\{ \underset{ 1\leq m \leq n_k, 1\leq m' \leq n_{k'}}\min d(M_{k,m},M_{k',m'})\bigg\}}{\underset{1\leq k \leq K}\max \bigg\{\underset{1\leq m,m' \leq n_k}\max d(M_{k,m},M_{k,m'})\bigg\}},
	\end{equation}
	which lies between 0 and $\infty$ with a higher value indicating better clustering.\\
	
	Connectivity:
	If the $j-$th nearest member of $M_{k,m}$ belongs to $C_k$, then we define a quantity $I_{k,m}(j)=0$, otherwise $I_{k,m}(j)=1/j$; where proximity between members is measured by distance. The connectivity (Handl et al. 2005) is given by
	\begin{equation}
		\text{Conn}=\sum\limits_{k=1}^{K}\sum\limits_{m=1}^{n_k}\sum\limits_{j=1}^{J}I_{k,m}(j), 
	\end{equation}
	where the value of the parameter $J$ is chosen to be 10 in our study. This measure takes a non-negative value where a lower value indicates better clustering.\\
	
	Cali\'{n}ski and Harabasz index:
	Let the mean of all members belonging to the cluster $C_k$ be denoted by $\overline{M}_{k,0}$ and the grand mean of all members of the data set be represented by $\overline{M}_{0,0}$, then 
	$\overline{M}_{k,0}=\frac{1}{n_k}\sum\limits_{m=1}^{n_k}M_{k,m}$ and $\overline{M}_{0,0}=\frac{1}{n}\sum\limits_{k=1}^{K}n_k\overline{M}_{k,0}$. The Cali\'{n}ski and Harabasz index (Cali\'{n}ski \& Harabasz 1974) is designed as 
	\begin{equation}
		\text{CH}=\frac{\sum\limits_{k=1}^{K}n_k d(\overline{M}_{k,0}\large{,}\overline{M}_{0,0})/(K-1)}{\sum\limits_{k=1}^{K}\sum\limits_{m=1}^{n_k}d(M_{m,k}\large{,}\overline{M}_{k,0})/(n-K)}, 
	\end{equation}
	which lies between 0 and $\infty$. It should be maximized for the best possible clustering.\\
	
	Nearest neighbor classification error rate:
	Nearest neighbor method is used to validate a classification scheme (Ripley 1996). Here for each clustered member $M_{k,m}$, we find its $l$ nearest members or neighbors (NNs) in terms of distance and consider their respective clusters. If the majority of the $l$ NNs belong to $C_k$, then we declare membership of $M_{k,m}$ to cluster $C_k$ acceptable and record a quantity $I_l(k,m)=0$, otherwise $I_l(k,m)=1$. If there is a tie by $l$ NNs, then we consider $I_l(k,m)=0$ or $1$ randomly. The required nearest neighbor classification error rate is defined by 
	\begin{equation}
		\text{NNCER}=\frac{1}{n}\sum\limits_{k=1}^{K}\sum\limits_{m=1}^{n_k}I_l(k,m), 
	\end{equation}
	where the value of the parameter $l$ is chosen to be 10 in our study. NNCER can have a value from 0 to 1 and should be minimized for the desired clustering.
	
	\subsection{Simulation}
	To asses the efficacy of our measure to identify the natural groups existing in a given data set and to compare its performance with its rival measures, we perform an extensive simulation study where samples are drawn from different groups and mixed together to be considered as one data set. Then cluster analysis is carried out on that data set with the help of a clustering algorithm for different number of clusters $K=2,3,\ldots$ and our clustering accuracy measure R$_{clus}$ along with its four competitors are computed for each $K$. The number of clusters for which a measure attains its optimal value, e.g. R$_{clus}$ reaches its highest value, is determined as the possible number of groups present in the data set which is to be compared with the true number of groups. For clustering we consider the Hartigan--Wong $K-$means method (Hartigan \& Wong 1979), agglomerative hierarchical algorithms (Kaufman \& Rousseeuw 2005) and DBSCAN clustering algorithm (Ester et al. 1996; Campello et al. 2013). It is to be noted that DBSCAN does not need to be run for different values of $K$ as it determines the value itself. The Euclidean metric is used here as the distance measure. 
	
\begin{figure}
		\centering
		\includegraphics[width=1\textwidth]{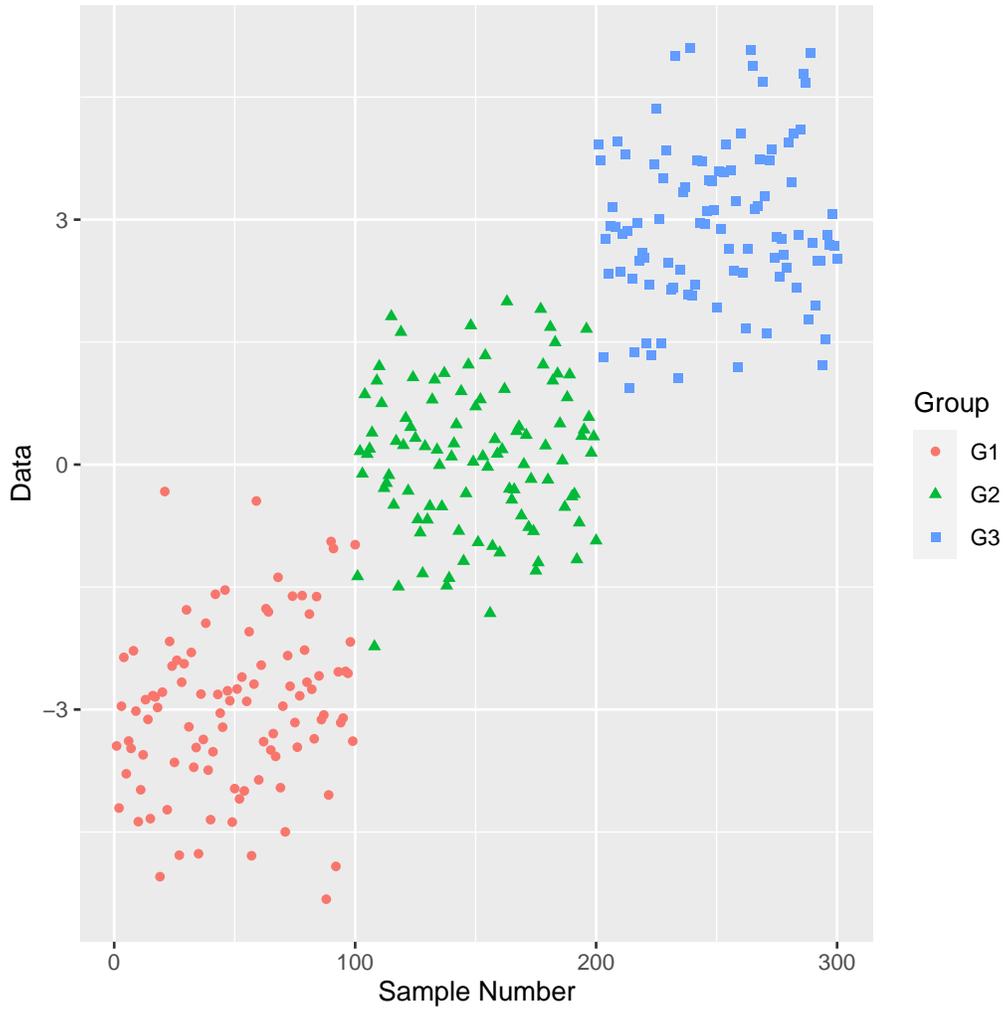}
		\caption{Univariate data set having three groups $G1,G2,G3$ each consisting of a random sample of size 100 independently drawn from the three populations $N(-3,1),N(0,1)$ and $N(3,1)$, respectively.}\label{UnivariateSimulation}
	\end{figure}

	1) First we consider univariate normal population with mean $\mu$ and standard deviation $\sigma$ denoted by $N(\mu,\sigma)$. Random samples of sizes 100 are drawn from each of the three normal populations $N(-3,1),N(0,1)$ and $N(3,1)$ independently. We mix the samples to form our data set of three narrowly separated homogeneous groups of spherical shapes (Fig.~\ref{UnivariateSimulation}). We perform $K-$means clustering on the data with $K=2,\ldots,6$ and compute our clustering accuracy measure R$_{clus}$, which attains its maximum at $K=3$ (Table~\ref{t:UnivariateSimulation}). For ease of reference, the true value of $K$ and the optimal values of different validity measures to estimate $K$ are highlighted in bold type (see, Table~\ref{t:UnivariateSimulation}). This clearly shows that our accuracy measure is capable of identifying natural clusters in a given data set. In this case, the inherent clustering structure is successfully revealed by our validity index, whereas its competitors fail to do so. 
	
\begin{figure}
		\centering
		\includegraphics[width=1\textwidth]{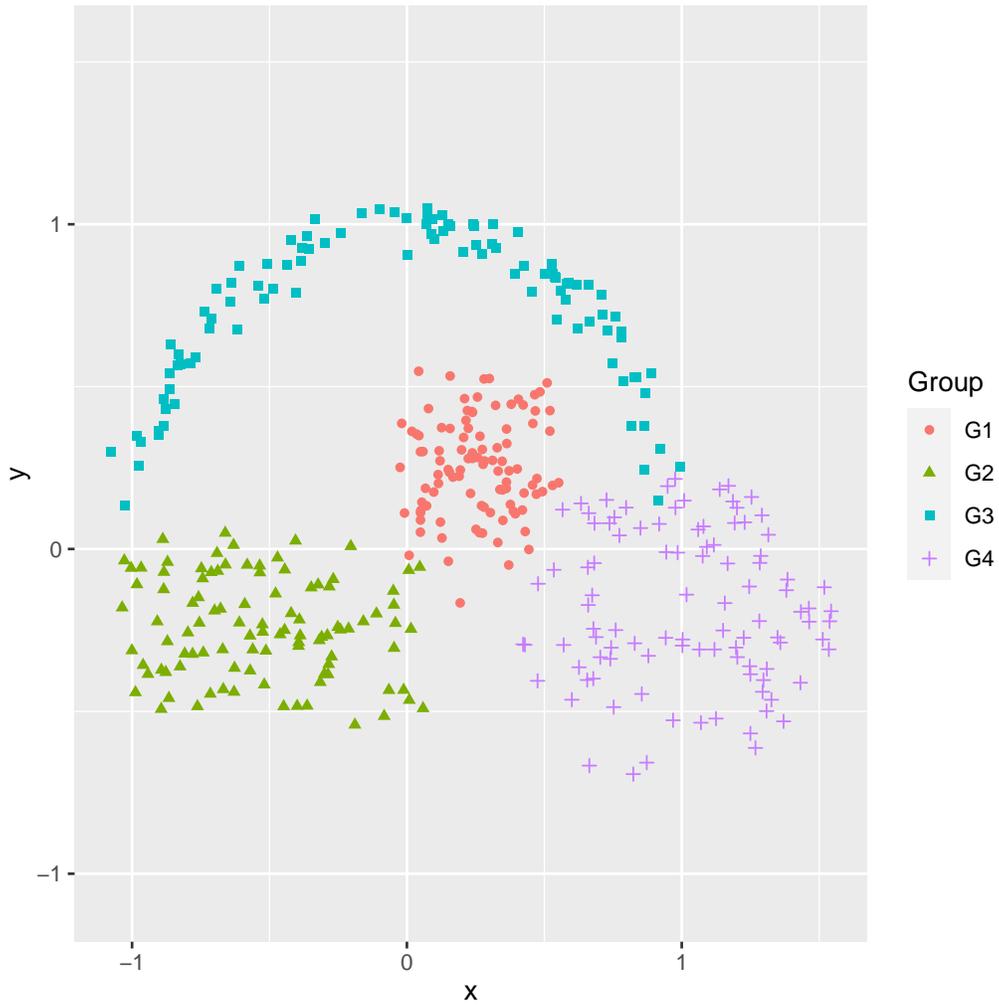}
		\caption{Bivariate data set having four groups $G1,G2,G3,G4$ each consisting of a random sample of size 100 independently drawn from a square, rectangle, curve and circle shaped clusters, respectively, with added Gaussian noise (0, 0.05).}\label{DiffshapedSimu}
	\end{figure}

	2) Secondly, we study a bivariate data set having noisy observations from four differently shaped groups (Fig.~\ref{DiffshapedSimu}). We draw samples of sizes 100 from each group of arbitrary shape, namely square, rectangle, curve and circle; and contaminate the observations variable-wise by adding Gaussian noise with mean 0 and standard deviation 0.05. Here our measure in association with $K-$means algorithm successfully reveals the true $K$ as 4, while the other measures could not (Table~\ref{t:DiffshapedSimu}).
	
	To check the robustness with respect to a different clustering method, we apply the DBSCAN algorithm (Ester et al. 1996; Campello et al. 2013), a clustering method that is known to  efficiently identify arbitrary shaped clusters. Unlike $K-$means, this approach does not need the number of clusters to be specified as a priori, although it depends upon two parameters `$\epsilon$' and `$Minpts$' to form the clusters. The members lying outside the $\epsilon-$neighborhood of the clustered data are considered as noise. The computed values for the clustering measures corresponding to different values of the parameters are as follows: For $\epsilon=0.140$ and $Minpts = 5$ we have (R$_{clus}\times 10^2$ = 21.940, Dunn$\times 10^2$ = 3.746, Conn = 4.050, CH = 84.639, NNCER = 0.254$\%$), whereas $\epsilon=0.165$ and $Minpts = 10$ produce (R$_{clus}\times 10^2$ = 24.792, Dunn$\times 10^2$ = 2.397, Conn = 16.008, CH = 65.162, NNCER = 1.036$\%$). In the first set-up, DBSCAN indicates wrongly two clusters with seven noise members while in the latter case it successfully exposes the true number of groups as four with 14 noise members. Therefore, it is clear that solely our measure can help to take the correct decision in this situation. Moreover, it does so, robustly, with both the clustering algorithms $K-$means and DBSCAN.     
	
	3) We simulate a high-dimensional data set which consists of groups of sizes 20, 15 and 10 drawn from three different 100-variate normal populations with mean vectors having all entries equal to $0,-3$ and 3, respectively. Here the multivariate dependence structure is constructed using a $t-$copula characterized by the $100-$variate $t-$distribution with $2$ degrees of freedom and correlation matrix having all the off--diagonal entries equal to $0.15$ (Nelsen 2006; Modak \& Bandyopadhyay 2019). Let $\textbf{T}$ denote the distribution function of the $100-$variate $t-$distribution where $\textbf{T}_i$ be the marginal distribution function for the $i-$th variable with inverse function $\textbf{T}_i^{-1}$, then the $t-$copula is expressed as
	\begin{equation*}
		C(u_1, . . . , u_{100}) = \textbf{T}\{\textbf{T}_1^{-1} (u_1), . . . , \textbf{T}_{100}^{-1}(u_{100})\}, 0<u_1,\ldots,u_{100}<1.
	\end{equation*} Results of $K-$means clustering of these data are shown in Table~\ref{t:CopulaHighdim}, where R$_{clus}$ is computed for different values of $w$ to check if the value of $w$ changes the outcome. It shows our measure consistently hints at the true $K=3$ for all considered values of $w$. On the other hand, the rest of the indices, except CH, do not reach the correct decision.
	
	Also, to investigate the stability of our validity index regarding different clustering algorithms and to show its efficacy to compare the results from various cluster analyses, we consider two different agglomerative hierarchical methods as follows. We perform hierarchical clustering with average linkage which classifies the data 100\% successfully into three clusters (the same as  $K-$means clustering does) for R$_{clus}=58.638$, whereas hierarchical algorithm with single linkage also hints at the optimal $K$ as 3 but for R$_{clus}= 28.115$ and merely with 46.666\% correct classification rate (here $w=0.1$). It justifies the fact that a higher value of R$_{clus}$ is an indicator of better clustering and this can robustly be used to compare the performances of different clustering methods when applied to the same data set.

	\subsection{Real data sets}
	
	1) First application is demonstrated through the worldwide ${CO}_2$ emission (metric tons per capita) data for the year 2011 over 199 countries, which are retrieved by Matioli et al. 2018 from the World Bank website \footnote{http://data.worldbank.org/indicator/EN.ATM.CO2E.PC/countries} early in the year 2016. Matioli et al. 2018 use a new nonparametric kernel density-based clustering method for univariate data and expose five existing clusters of the ${CO}_2$ emission data by the average silhouette width (Rousseeuw 1987). Here we show that $K-$means algorithm is efficient enough to confirm the same but only by our accuracy measure R$_{clus}$ with resulting five clusters of sizes 109, 56, 21, 11 and 2 (Table~\ref{t:Co2}). The optimal values of different measures to estimate $K$ are highlighted in bold type in the Table.
	
\begin{figure}
		\centering
		\includegraphics[width=1\textwidth]{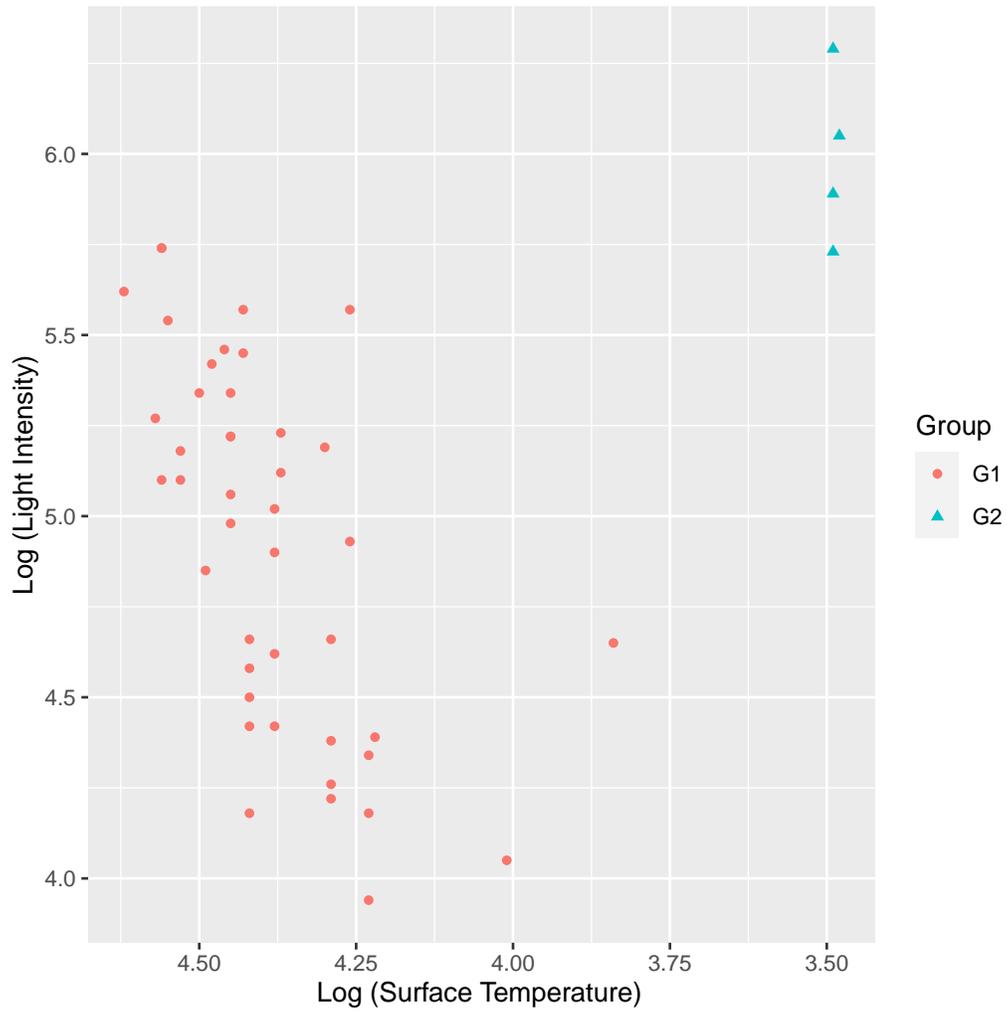}
		\caption{The Hertzsprung-Russell diagram of the `CYG OB1' star cluster data showing group $G1$ of 43 stars included in the main sequence and $G2$ presents the group of 4 giant stars.}\label{CYG-OB1}
	\end{figure}

	2) Secondly, we choose a bivariate astronomical data set on a star cluster `CYG OB1' known to consist of two different inherent groups (Vanisma \& Greve 1972; Kaufman \& Rousseeuw 2005). The group of 43 stars belonging to main sequence is well separated from the other group of 4 giant stars. Fig.~\ref{CYG-OB1} shows the Hertzsprung-Russell diagram of the logarithm of surface temperature as a study variable against the logarithm of light intensity as the other study variable for all the 47 stars. As the two groups appear to be elongated in $2-$dimensional plane, we apply agglomerative hierarchical clustering with single linkage and expose the original two clusters in terms of our validity index R$_{clus}$ (Table~\ref{t:CYG-OB1}). These widely separated clusters are acknowledged by all the other indices under comparison.
	
	3) Our third data set comprises the gamma-ray bursts (GRBs)  that are the brightest source of light in the Universe after the Big Bang (Bandyopadhyay \& Modak 2018; Modak et al. 2018; Modak 2021). Cluster analysis of GRBs is an ongoing vital task to confirm whether two or three groups are required to describe their cosmological origins (Norris et al. 1984; Kouveliotou et al. 1993; Mukherjee et al. 1998; Tarnopolski 2015; Modak et al. 2018; Toth et al. 2019; Modak 2021). We consider the current BATSE Gamma-Ray Burst Catalog\footnote{https://gammaray.nsstc.nasa.gov/batse/grb/catalog/current/} (Toth et al. 2019; Modak 2021) of 1,972 GRBs for the following observed variables: fluences $F_1,F_2,F_3,F_4$, peak fluxes $P_{64},P_{256},P_{1024}$ and durations $T_{50},T_{90}$.  Modak et al. (2018) perform a new  machine learning method, that is, the kernel principal component analysis (Sch\"{o}lkopf \& Smola 2002; Modak et al. 2017) on the standardized forms of the above--mentioned variables. We choose the same study variables as Modak et al. (2018), which are the first two kernel principal components extracted through a novel kernel, i.e. their kernel (10) with hyperparameters $s=\sigma_1=0.937$ and $p=1/2$. Modak et al. (2018) show that $K-$means clustering based on these variables gives three clusters indicated by the gap statistic (Tibsirani et al. 2001), which produce the optimal classification of the data as validated by the Dunn index (see, their Table 2). Here also, $K-$means clustering gives the same partitions in terms of our accuracy measure and the Dunn index (see, our Table~\ref{t:GRB}), which supports the existence of three statistically significant clusters in the GRB population (Mukherjee et al. 1998; Balastegui et al. 2001; Chattopadhyay et al. 2007; King et
	al. 2007; Veres et al. 2010; Horv\'{a}th et al. 2018; Modak et al. 2018; Toth et al. 2019; Modak 2021). However, the connectivity measure estimates $K=2$ and NNCER holds the ambiguity between two and three clusters; whereas CH is observed to be increasing in value as $K$ increases and thus proved to be worthless for these data. 
	
	\section{Conclusion}
	In this paper, we propose a new nonparametric, interpoint distance-based cluster validity index. This assessment criterion is compatible with any distance measure and clustering algorithm, which can be applied to univariate and multivariate data sets having observations measured on arbitrary scales. Synthetic and real-life data study establishes its superiority to other competitors in identifying the true number of clusters existing in the data sets by assessing the quality of a cluster analysis, and helping with the selection of the best possible clustering of the data members by comparing the relative performances of different clustering algorithms. Wide applicability, high-dimensional use, simple computation, easy interpretation and great performance ascertain that our novel index is a very useful measure for assessment of clustering. 
	
	\section{Acknowledgements}
	The author would like to thank the Editors for encouraging
	the present work. The author expresses sincere gratitude to one anonymous referee for valuable advice and intriguing inquiries 
	which helped the author to present the paper in a more convincing way.
	\clearpage
	\begin{table}
		\caption{Computed values of various measures for different number of clusters ($K$) as  obtained by $K-$means clustering of the univariate data set containing three inherent groups from normal populations}
		\begin{center}
			\begin{tabular}{cccccc }
				\hline\\
				$K$   & R$_{clus}$&Dunn&Conn&CH&NNCER\\[1ex]
				& ($\times 10^{2}$) &($\times 10^{2}$)&&&(\%)\\[1ex]
				\hline\\
				
				2&61.989& 0.897&   \textbf{4.318}&  803.311& \textbf{0.000}\\[1ex]
				\textbf{3}&\textbf{70.503}& 1.070& 11.802& 1215.793& 1.000\\[1ex]
				4&68.500& \textbf{2.290}& 9.986& 1281.206& 1.000\\[1ex]
				5&66.175& 0.709& 19.136& 1271.142& 2.333\\[1ex]
				6&67.440& 1.988& 22.827& \textbf{1695.936}& 2.333\\[1ex]
				
				\hline
			\end{tabular}
		\end{center}
		\label{t:UnivariateSimulation}
	\end{table}
	\clearpage
	\begin{table}
		\caption{Computed values of various measures for different number of clusters ($K$) as  obtained by $K-$means clustering of the bivariate data set from four differently shaped noisy groups}
		\begin{center}
			\begin{tabular}{cccccc }
				\hline\\
				$K$   & R$_{clus}$&Dunn&Conn&CH&NNCER\\[1ex]
				& ($\times 10^{2}$) &($\times 10^{2}$)&&&(\%)\\[1ex]
				\hline\\
				2&64.847& 2.043& \textbf{19.198}& 386.458& 1.25\\[1ex]
				3&63.415& 3.916& 19.897& 490.960& \textbf{0.25}\\[1ex]
				\textbf{4}&\textbf{67.305}& 3.243& 24.438& 508.885& 1.00\\[1ex]
				5&64.025&\textbf{ 4.225}& 42.340& 463.659& 2.50\\[1ex]
				6&64.570& 3.369& 51.837& \textbf{512.928}& 4.25\\[1ex]
				\hline
			\end{tabular}
		\end{center}
		\label{t:DiffshapedSimu}
	\end{table}
	\clearpage
	\begin{table}
		\caption{Computed values of various measures for different number of clusters ($K$) as  obtained by $K-$means clustering of the high-dimensional data from three normal populations with multivariate structure specified by a $t-$ copula}
		\begin{center}
			\begin{tabular}{cccccccc }
				\hline\\
				$K$ & R$_{clus}$ ($\times 10^{2}$)&R$_{clus}$ ($\times 10^{2}$)&R$_{clus}$ ($\times 10^{2}$)&Dunn&Conn&CH&NNCER\\
				&  $w=0.025$ &$w=0.05$ & $w=0.1$ &($\times 10^{2}$)     &  & &(\%)\\[1ex]
				\hline\\
				
				2&39.721&46.026&49.168& \textbf{53.303}&\textbf{1.236}&51.385&  \textbf{0}\\[1ex]
				\textbf{3}&\textbf{47.364}&\textbf{56.848}&\textbf{58.638}& 49.609&  1.901& \textbf{83.184}&\textbf{0}\\[1ex]
				4&14.791&16.384&18.056& 36.940& 10.577& 57.935&  6.667\\[1ex]
				5&15.763&16.649&20.409& 36.940& 13.506& 45.947&  8.889		\\[1ex]
				6&14.592&15.626&18.226& 36.940& 16.434& 38.725& 11.111		\\[1ex]
				
				\hline
			\end{tabular}
		\end{center}
		\label{t:CopulaHighdim}
	\end{table}
	\clearpage
	\begin{table}
		\caption{Computed values of various measures for different number of clusters ($K$) as  obtained by $K-$means clustering of the $CO_2$ emission data}
		\begin{center}
			\begin{tabular}{cccccc }
				\hline\\
				$K$   & R$_{clus}$&Dunn&Conn&CH&NNCER\\[1ex]
				& ($\times 10^{2}$) &($\times 10^{2}$)&&&(\%)\\[1ex]
				\hline\\
				2&23.742& \textbf{2.477}&\textbf{4.169}& 333.603& \textbf{0.503}\\[1ex]
				3&31.443& 0.355& 13.569& 422.361& 2.010\\[1ex]
				4&55.143& 0.629& 16.440& 625.837& 2.513\\[1ex]
				5&\textbf{71.763}& 2.439& 18.325& 839.647& 2.513\\[1ex]
				6&58.619& 0.813& 20.000& \textbf{853.928}& 3.518\\[1ex]

				\hline
			\end{tabular}
		\end{center}
		\label{t:Co2}
	\end{table}
	
	\clearpage
	\begin{table}
		\caption{Computed values of various measures for different number of clusters ($K$) as  obtained by hierarchical clustering of the data from `CYG OB1' star cluster}
		\begin{center}
			\begin{tabular}{cccccc }
				\hline\\
				$K$   & R$_{clus}$&Dunn&Conn&CH&NNCER\\[1ex]
				& ($\times 10^{2}$) &($\times 10^{2}$)&&&(\%)\\[1ex]
				\hline\\
				2&\textbf{72.090}&\textbf{42.975}& \textbf{ 4.383}& \textbf{27.308}&\textbf{8.511}\\[1ex]
				3&36.702& 24.596&  7.312& 14.534& 10.638\\[1ex]
				4&37.624& 13.441& 10.865& 11.623& 12.766\\[1ex]
				5&33.100& 13.126& 13.865&  8.736& 12.766\\[1ex]
				6&13.983& 15.052& 16.760&  8.572& 14.894\\[1ex]
				
				\hline
			\end{tabular}
		\end{center}
		\label{t:CYG-OB1}
	\end{table}
	\clearpage
	\begin{table}
		\caption{Computed values of various measures for different number of clusters ($K$) as  obtained by $K-$means clustering of the GRB data set}
		\begin{center}
			\begin{tabular}{ccccccc }
				\hline\\
				$K$   & R$_{clus}$ &Dunn&Conn&CH&NNCER\\[1ex]
				& ($\times 10^{2}$) &($\times 10^{2}$)&&&(\%)\\[1ex]
				
				\hline\\
				2 &     78.601& 0.612& \textbf{16.173}& 2734.160&\textbf{ 0.203}\\[1ex]
				3&\textbf{81.427}&\textbf{1.885}& 24.887&6040.294&\textbf{0.203}\\[1ex]
				4      &    69.569& 0.384& 69.482& 6596.430& 0.862\\[1ex]
				5      &    70.805& 0.566& 77.259& 7945.977& 0.659\\[1ex]
				6      &    68.851& 0.644& 79.102& \textbf{8558.447}& 0.913\\[1ex]

				\hline
			\end{tabular}
		\end{center}
		\label{t:GRB}
	\end{table}
	
	\clearpage
	\clearpage
	\clearpage
	{}
\end{document}